\documentclass{article}

\usepackage{times}
\usepackage{graphicx} % more modern
\usepackage{comment}
\usepackage{lipsum}
\usepackage{color}
\usepackage[colorinlistoftodos]{todonotes}

\usepackage{algorithm}
\usepackage{algorithmic}
\usepackage{amsmath}
\usepackage{amssymb}
\usepackage{amsfonts}
\usepackage{amsthm}
\usepackage{mathtools}
\usepackage{bm}
\usepackage{mathrsfs}
\usepackage{eucal} % makes mathcal standard and CMcal the flourished one

\usepackage{graphicx}
\usepackage{epstopdf}
\usepackage[font=footnotesize]{caption}
\usepackage{subfig}
\usepackage{wrapfig}
\usepackage{tikz}
\usepackage{tabularx}
\usepackage{placeins}
\usepackage{dblfloatfix}

\usepackage{booktabs}
\usepackage{float}
\usepackage{placeins}

\usepackage{hyperref}

\usepackage[accepted]{icml2017}

\icmltitlerunning{Transferring Autonomous Driving Knowledge on Simulated and Real Intersections}

\begin{document} 

\twocolumn[
\icmltitle{Transferring Autonomous Driving Knowledge on Simulated and Real Intersections}

\begin{icmlauthorlist}
\icmlauthor{David Isele}{upenn,honda}
\icmlauthor{Akansel Cosgun}{honda}
\end{icmlauthorlist}

\icmlaffiliation{upenn}{University of Pennsylvania, Philadelphia, USA}
\icmlaffiliation{honda}{Honda Research Institute, Mountain View, California, USA}

\icmlcorrespondingauthor{David Isele}{isele@seas.upenn.edu}
\icmlcorrespondingauthor{Akansel Cosgun}{akansel.cosgun@gmail.com}

\icmlkeywords{transfer, reinforcement learning, autonomous driving}

\vskip 0.3in
]



\printAffiliationsAndNotice{}  % leave blank if no need to mention equal contribution

\begin{abstract} 
We view intersection handling on autonomous vehicles as a reinforcement learning problem, and study its behavior in a transfer learning setting. We show that a network trained on one type of intersection generally is not able to generalize to other intersections. However, a network that is pre-trained on one intersection and fine-tuned on another performs better on the new task compared to training in isolation. This network also retains knowledge of the prior task, even though some forgetting occurs. Finally, we show that the benefits of fine-tuning hold when transferring simulated intersection handling knowledge to a real autonomous vehicle.
\end{abstract} 

\section{Introduction}

Autonomous Driving (AD) has the potential to reduce accidents caused by driver fatigue and distraction and will enable more active lifestyles for the elderly and disabled. While AD technology has made important strides over the last couple of years, current technology is still not ready for large scale roll-out. Urban environments are particularly difficult for AD, due to the unpredictable nature of pedestrians and vehicles in city traffic. 

Rule-based methods provide a predictable method to handle intersections. However, rule-based intersection handling approaches do not scale well due to the difficulty of designing hand-crafted rules that remain valid as the diversity and complexity of possible scenes increase. Recently it has been shown that deep reinforcement learning can improve over rule-based techniques \cite{isele2016task}, however it is unclear how well these techniques are able able to generalize to different scenarios and real systems. %We model the AD vehicle as a learning agent, which learns from positive (successful passing) and negative (collisions) experiences in a reinforcement learning framework.

\begin{figure}[t!]
\subfloat[Direct Copy\label{fig:transfer0}]{%
\includegraphics[width=.15\textwidth]{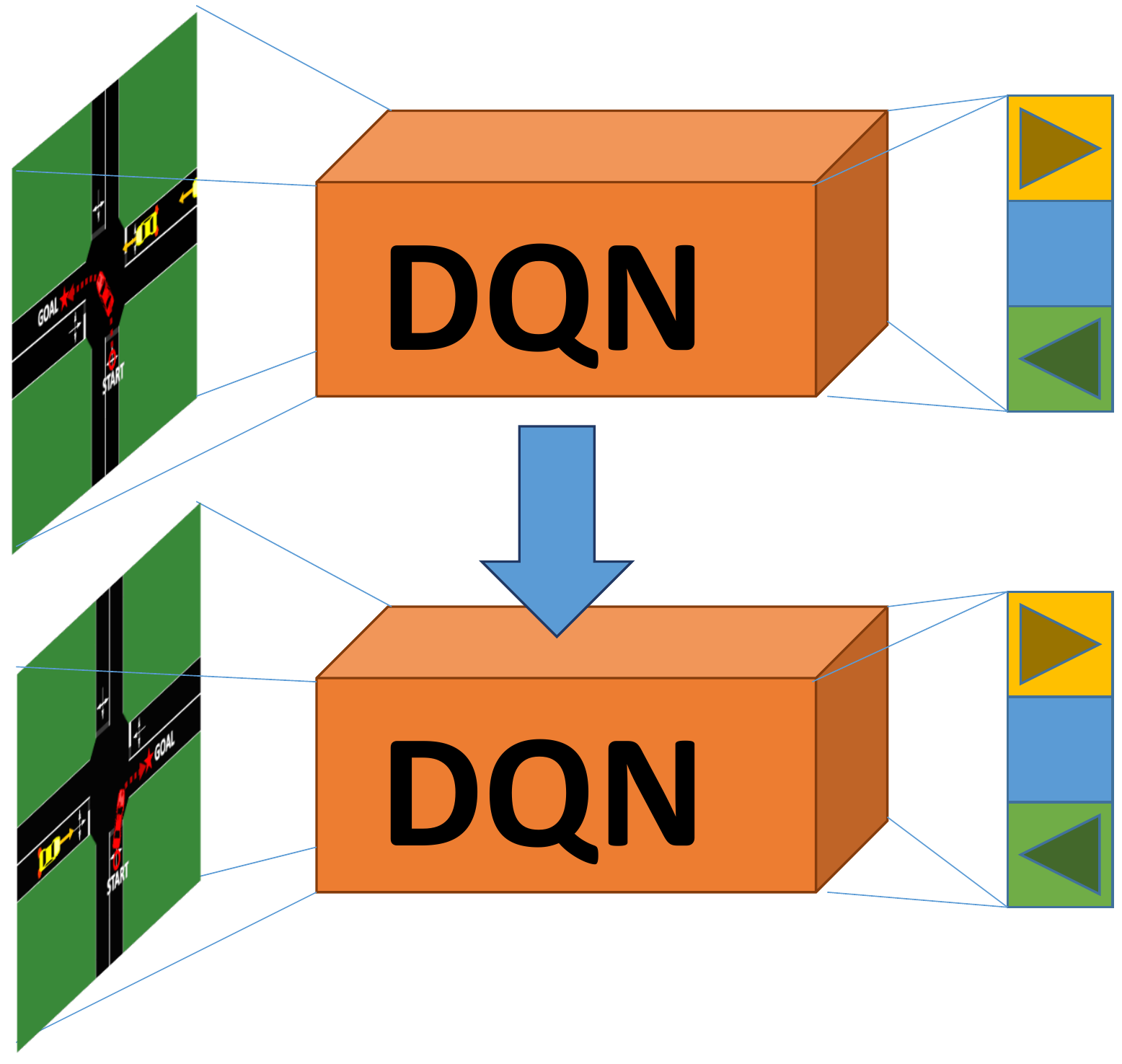}
}
%\hfill
\hspace{1pt}
\subfloat[Fine Tuning\label{fig:transfer1}]{%
\includegraphics[width=.15\textwidth]{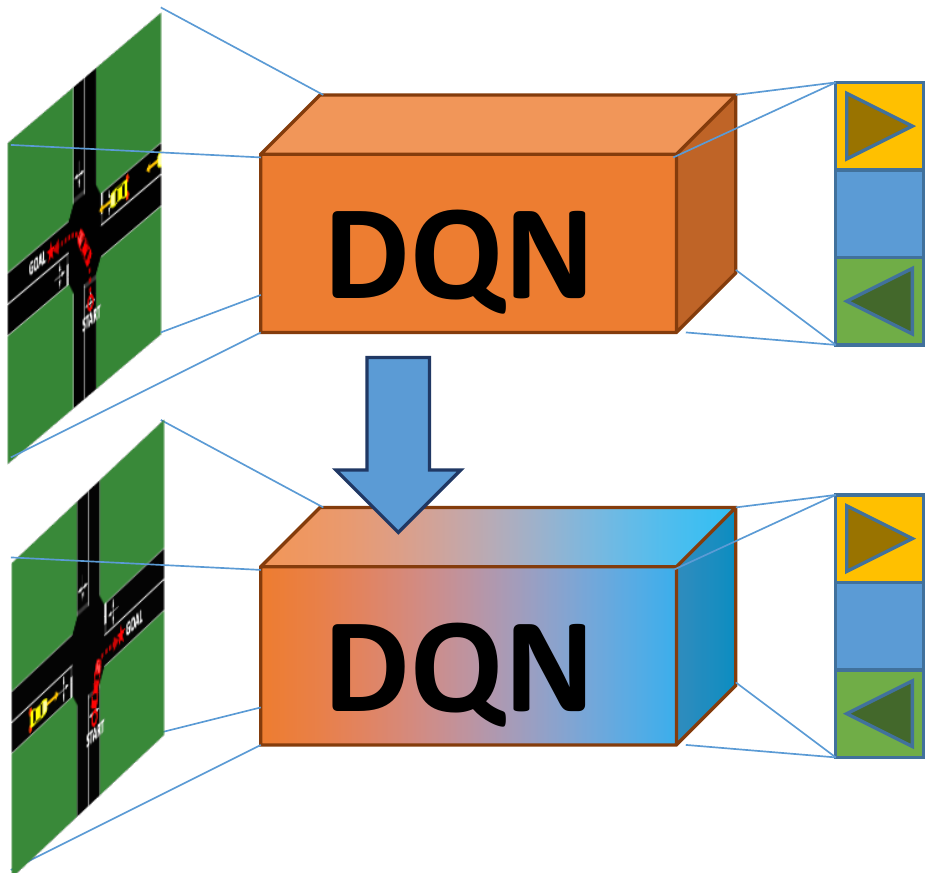}
}
%\hfill
\hspace{1pt}
\subfloat[Reverse Transfer\label{fig:transfer2}]{%
\includegraphics[width=.15\textwidth]{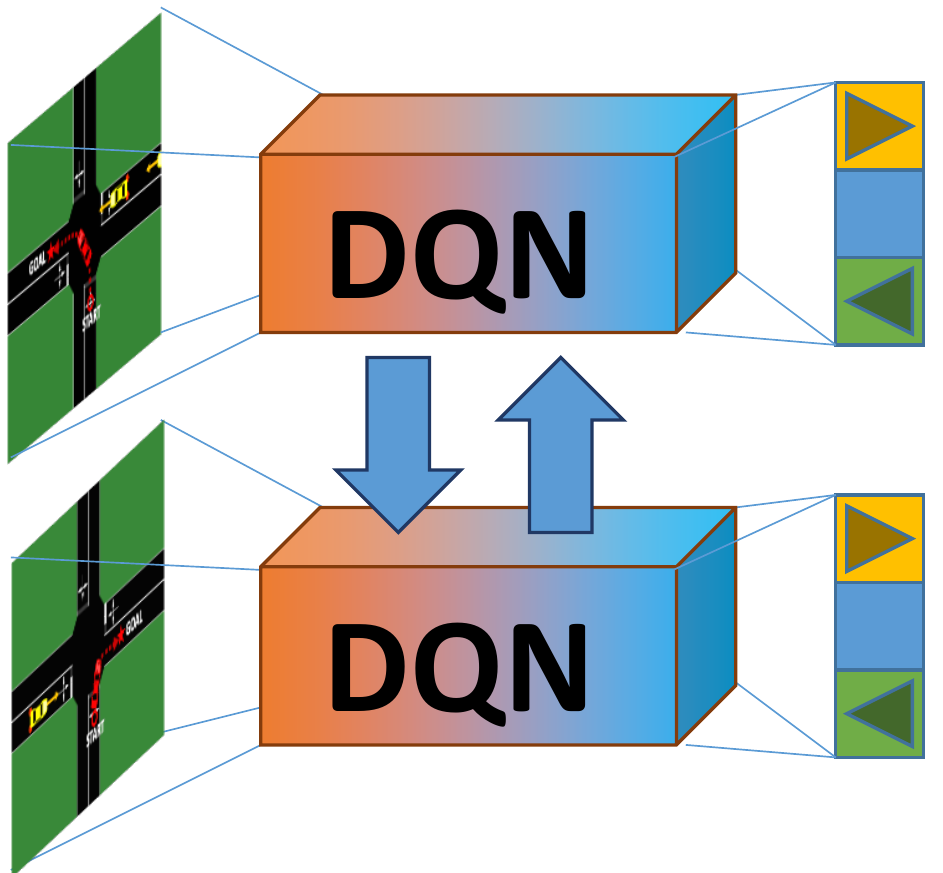}
}
\caption{We analyze knowledge transfer between different types of intersections. The knowledge to handle an intersection is represented as a Deep Q-Network (DQN). We investigate a) directly copying a network to a new intersection b) fine-tuning a previously trained network on a new intersection, c) whether fine tuning destroys old intersection knowledge in reverse transfer.}\label{fig:transfers}
\end{figure}

We explore the ability of a reinforcement learning agent to generalize,  focusing specifically on how the knowledge for one type of intersection, represented as a Deep Q-Network (DQN), translates to other types of intersections (tasks). First we look at \textbf{direct copy}: how well a network trained for Task A performs on Task B. Second, we analyze how a network initialized on Task A and \textbf{fine-tuned} on Task B compares to a randomly initialized network exclusively trained on Task B. Third, we investigate \textbf{reverse transfer}: if a network pre-trained for Task A and fine-tuned to Task B, preserves knowledge for Task A. Finally, we present early results of using a network trained in simulation to initialize learning on real data.

% This paper is organized as follows. After providing a brief literature survey in Section \ref{sec:related_works}, we present the problem formulation as a DQN in Section \ref{sec:intersection_handling}, before examining various knowledge sharing strategies in Section \ref{sec:knowledge}. After explaining the experimental setup in Section \ref{sec:experiments}, then present our results in Section \ref{sec:results} before concluding in Section \ref{sec:conclusion}.

\begin{figure*}[t!]
\subfloat[Right\label{fig:right}]{%
\includegraphics[height=.146\textwidth]{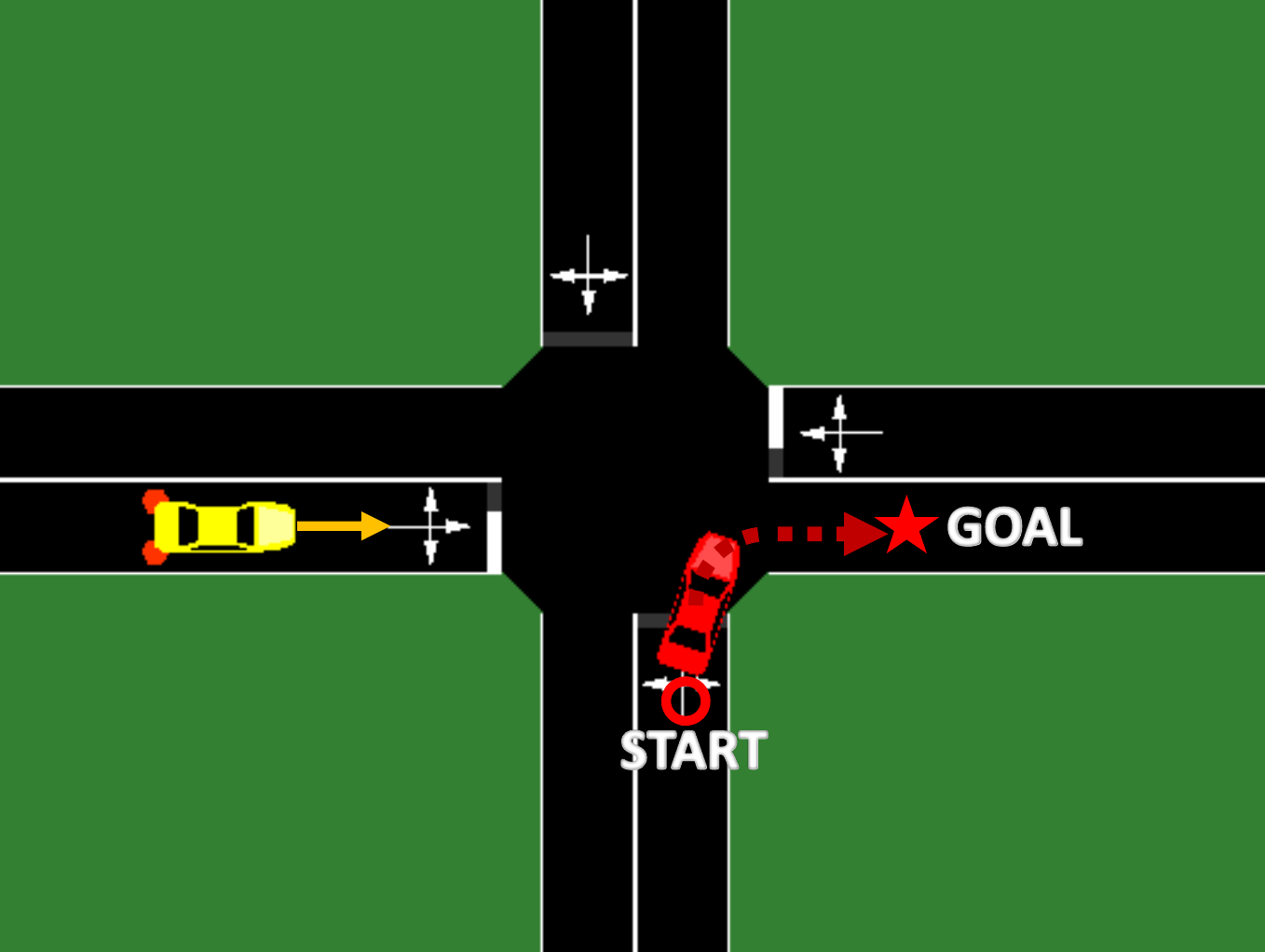}
}
\subfloat[Left\label{fig:left}]{%
\includegraphics[height=.146\textwidth]{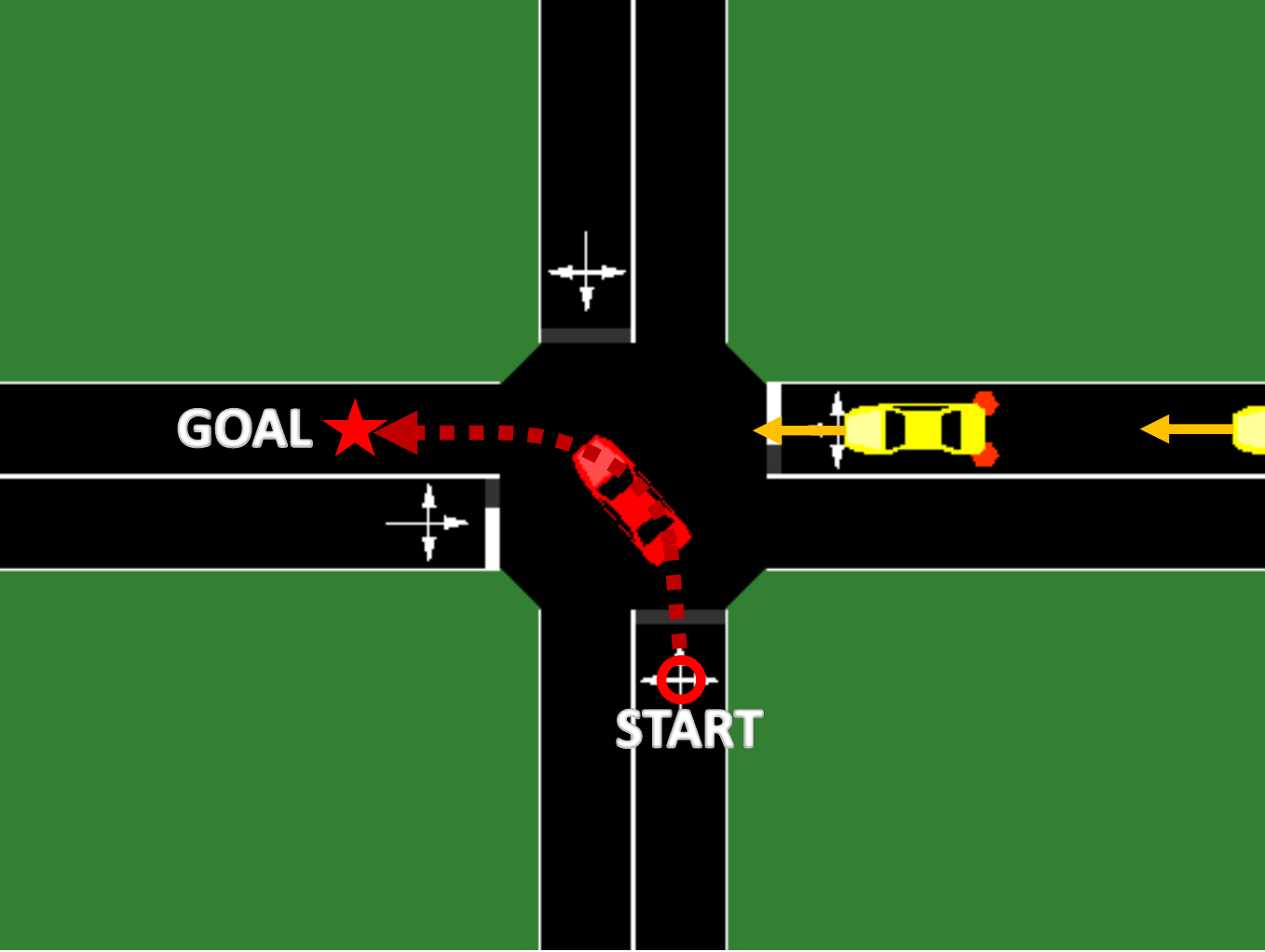}
}
\subfloat[Left2\label{fig:left2}]{%
\includegraphics[height=.146\textwidth]{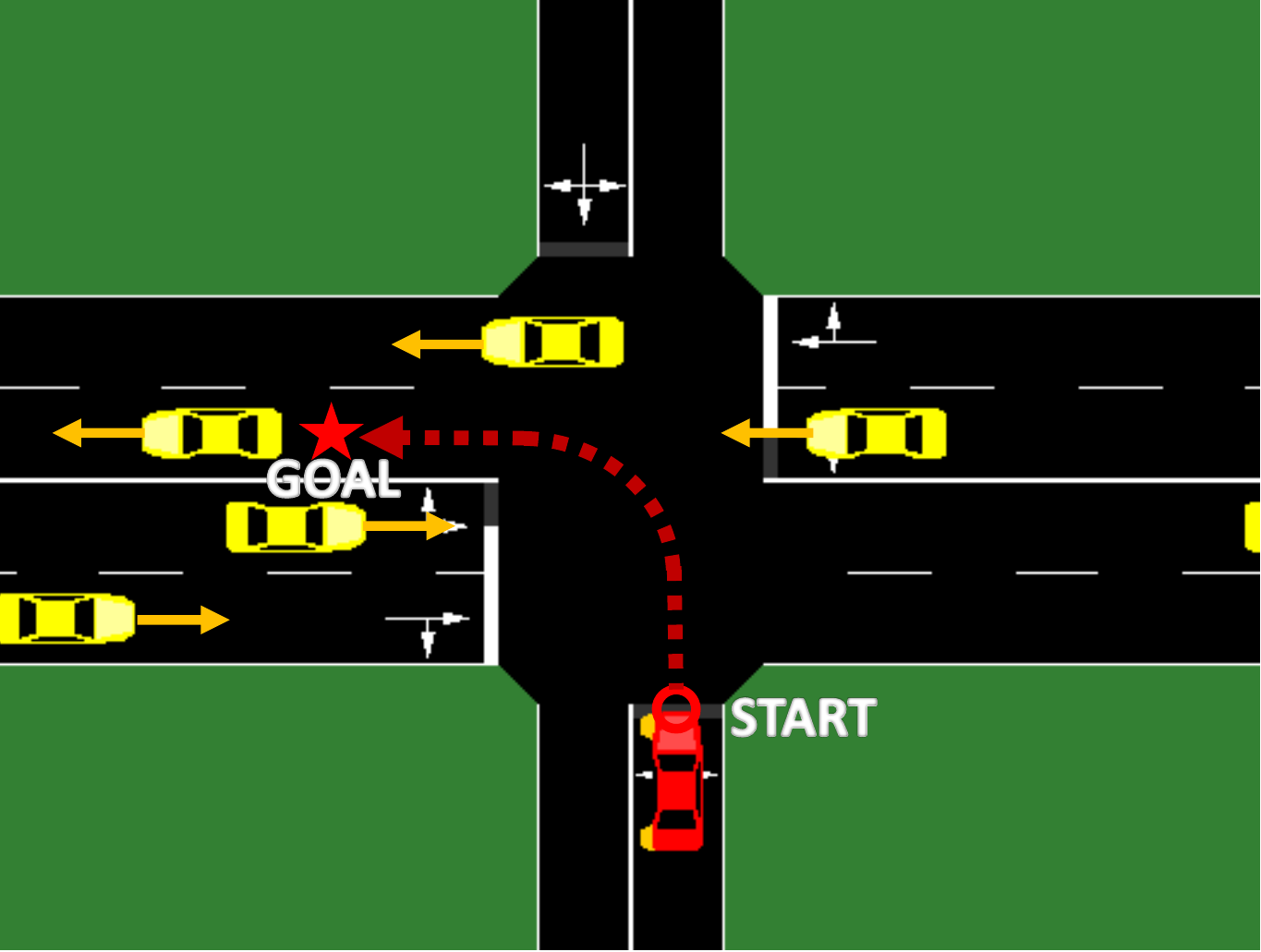}
}
\subfloat[Forward\label{fig:forward}]{%
\includegraphics[height=.146\textwidth]{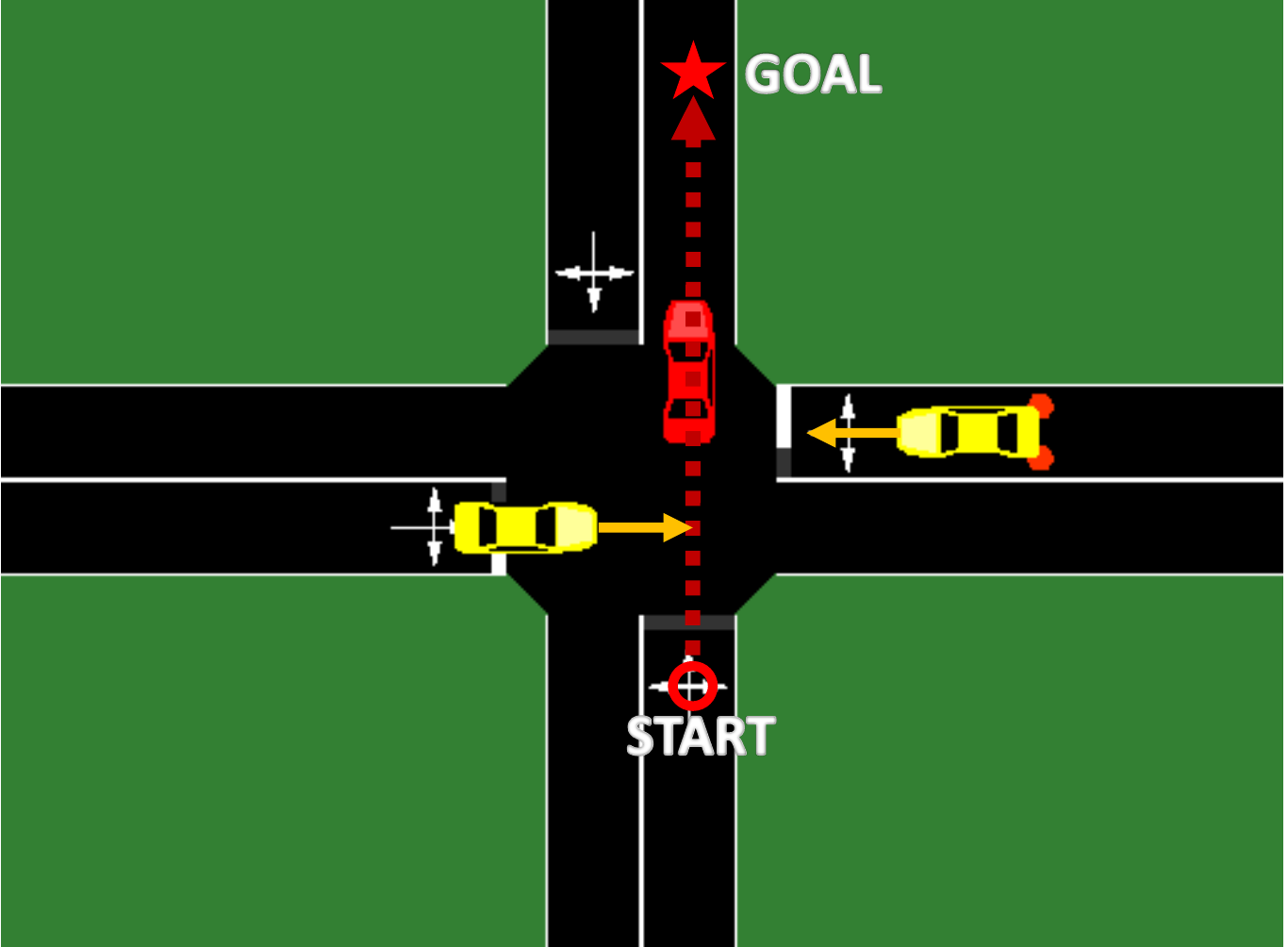}
}
\subfloat[Challenge\label{fig:challenge}]{%
\includegraphics[height=.146\textwidth]{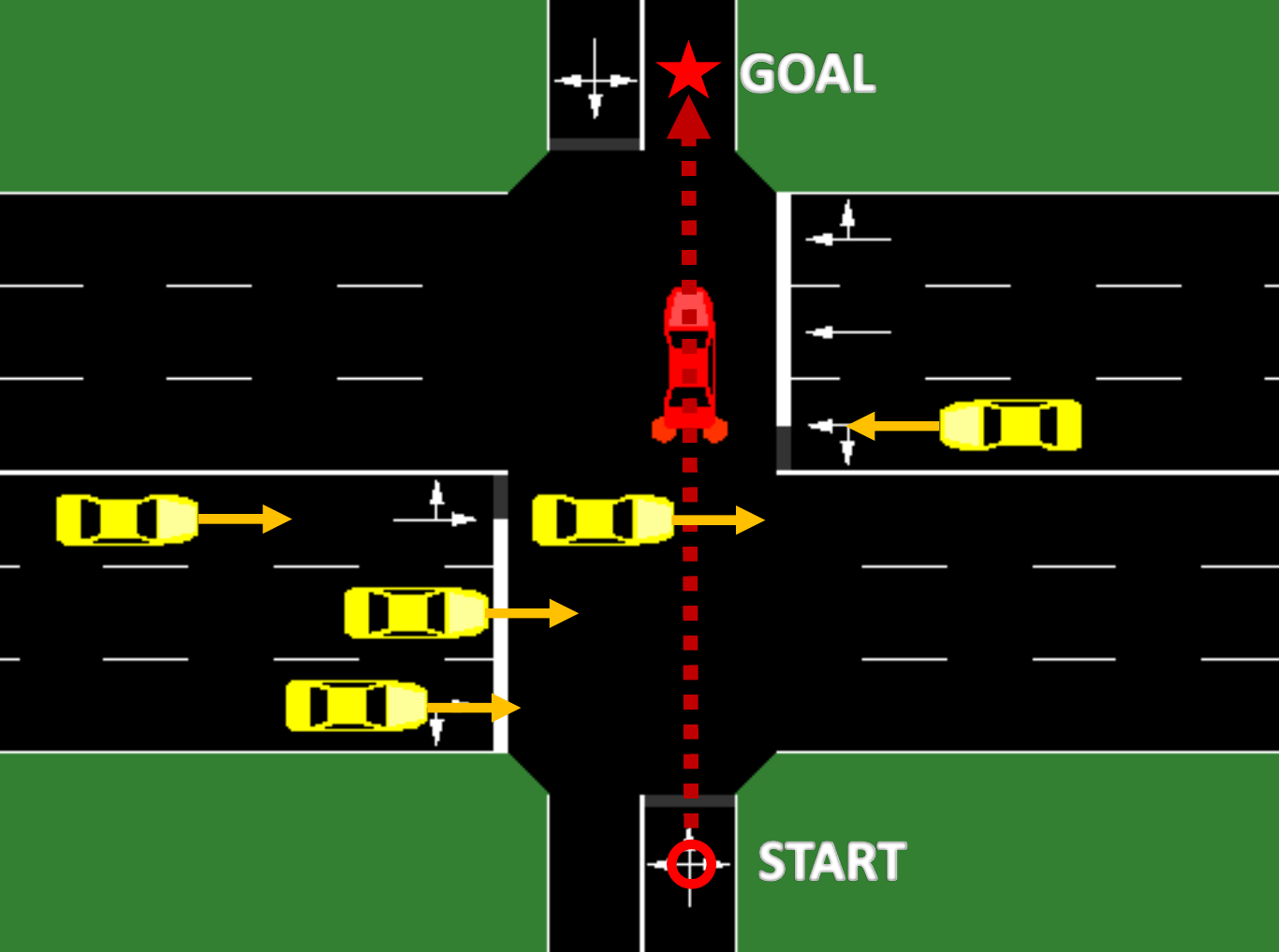}
}
\caption{Visualizations of different intersection scenarios.}\label{fig:scenarios}
\end{figure*}

\section{Related Work}
\label{sec:related_works}

% cars
Researchers have recently been investigating using machine learning techniques to control autonomous vehicles \cite{cosgun2017towards}.
Imitation learning strategies have investigated learning from a human driver \cite{bojarski2016end}.
%, but has difficulty if the agent finds itself in a state that is not part of the training data. 
Markov Decision Processes (MDP) have been used offline to address the problem of intersection handling \cite{brechtel2014probabilistic,song2016intention}. And 
online planners based on partially observable Monte Carlo Planning (POMCP) have been applied to intersection problems when an accurate generative model is available \cite{bouton2017belief}. Additionally, 
machine learning techniques have been used to optimize comfort in a space where solutions are constrained to safe trajectories \cite{shalev2016safe}.

% transfer 
Large amounts of data often improve the performance of machine learning techniques. %This helps a system learn general representations and prevents over fitting based on incidental correlations in the sampled data. 
In the absence of huge datasets, training on multiple related tasks can give similar performance gains \cite{Caruana1997}. A large breadth of research has investigated transferring knowledge from one system to another in machine learning in general \cite{Pan2010a}, and reinforcement learning specifically \cite{taylor2009transfer}. 

Large training times and high sample complexity make transfer methods particularly appealing in deep networks \cite{Razavian2014,Yosinski2014}. Recent work in deep reinforcement learning has looked at combining networks from different tasks to share information \cite{rusu2016progressive,yin2017knowledge}. Researchers have looked at using options \cite{sutton1998reinforcement} in Deep RL to expand an agent's capabilities
\cite{jaderberg2016reinforcement,tessler2016deep,kulkarni2016hierarchical},
and efforts have been made to enable a unified framework for learning multiple tasks through changes in architecture design \cite{srivastava2013compete} and modified objective functions \cite{kirkpatrick2016overcoming} to address the problem of catastrophic forgetting \cite{goodfellow2013empirical}. 

In our scenario, there is the added difficulty of transferring to the real vehicle. It is a well known problem that policies trained in simulation rarely work on real robots \cite{barrett2010transfer}. Recent work has investigated grounding imperfect simulators to a robot's behavior \cite{hanna2017grounded} and there is evidence that transferring from simulation to real robots can be addressed by targeting the systems ability to generalize \cite{tobin2017domain}. Given the variety of intersections, the importance of learning a general model, and the difficulty of training on the real vehicle, we examine the prospects of multi-task learning for the problem of intersection handling.

\section{Intersection Handling using Deep Networks}
\label{sec:intersection_handling}

Each intersection handling task is viewed as a reinforcement learning problem, and we use a Deep Q-Network (DQN) to learn the state-action value Q-function. We assume the vehicle is at the intersection, the path is known, and the network is tasked with choosing between two actions: wait or go, for every time step. Once the agent decides to go, it continues until it either collides or successfully navigates the intersection. Previous work has shown that deciding the wait time generally outperforms approaches that learn an entire acceleration profile \cite{isele2017navigating}. 

\subsection{Reinforcement Learning}

% RL framework
The reinforcement learning framework considers an agent in state $s$ taking an action $a$ according to the policy $\pi$. After taking an action, the agent transitions to the state $s'$% according to the environment $\mathcal{E}$
, and receives a reward $r$. This collection is defined as an experience  $e = (s,a,r,s')$. Learning is formulated as a Markov decision process (MDP) and follows the Markov assumption that the probability of transitioning to a new state given the current state and action is independent of all previous states and actions $p(s_{t+1}|s_{t}, a_{t}, \dots, s_{0}, a_{0}) = p(s_{t+1}|s_{t}, a_{t})$. 

The objective at time step $t$ is to maximize the future discounted return $R_t = \sum_{k=t}^T \gamma^{k-t} r_{k}$. We optimize this objective using Q-learning \cite{watkins1992q}. 

% Q LEARNING
\subsection{Q-Learning}
In Q-learning an optimal action-value function $Q^*(s,a)$ is defined as the maximum expected return that is achievable following any policy given a state $s$ and action $a$, $Q^*(s,a) = max_\pi \mathbb{E}[R_t|s_t = s, a_t = a, \pi]$. 

Deep Q-learning \cite{mnih2013playing} approximates the optimal value function with a neural network $Q^*(s,a)\approx Q(s,a;\theta)$. The parameters $\theta$ are learned by using the Bellman equation as an iterative update $Q_{i+1}(s,a) = \mathbb{E} [r + \gamma \max_{a'} Q_i(s',a')|s,a]$ and minimizing the error between the expected return and the state-action value predicted by the network. This gives the loss for an individual experience in a deep Q-network (DQN)   
%\vspace{-0.4cm}
\begin{eqnarray}
\mathcal{L}(e_i,\theta) = \bigg( r_i + \gamma \max_{a_i'}Q(s_i',a_i';\theta) - Q(s_i,a_i;\theta)\bigg)^2    \enspace .
\end{eqnarray}

\begin{figure*}[tbh!]
\subfloat[Right\label{fig:finetune_right}]{%
\includegraphics[height=.15\textwidth]{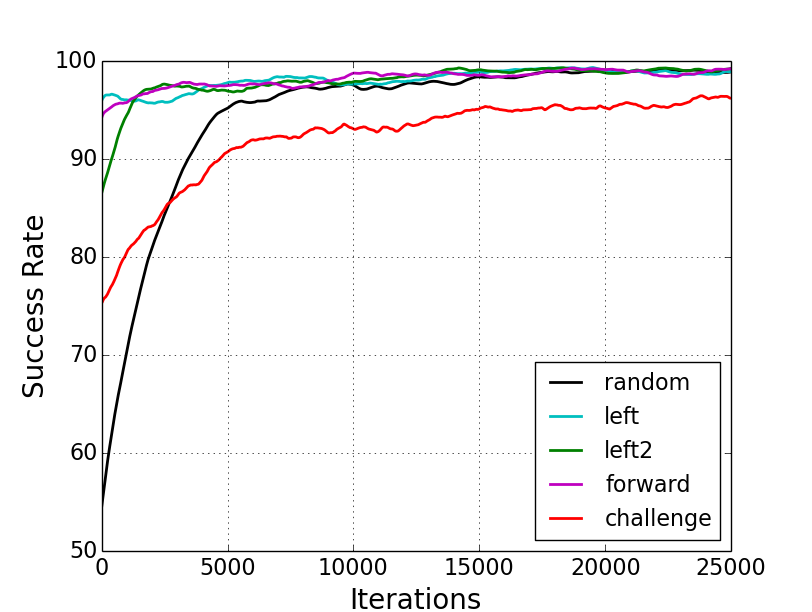}
}
\subfloat[Left\label{fig:finetune_left}]{%
\includegraphics[height=.15\textwidth]{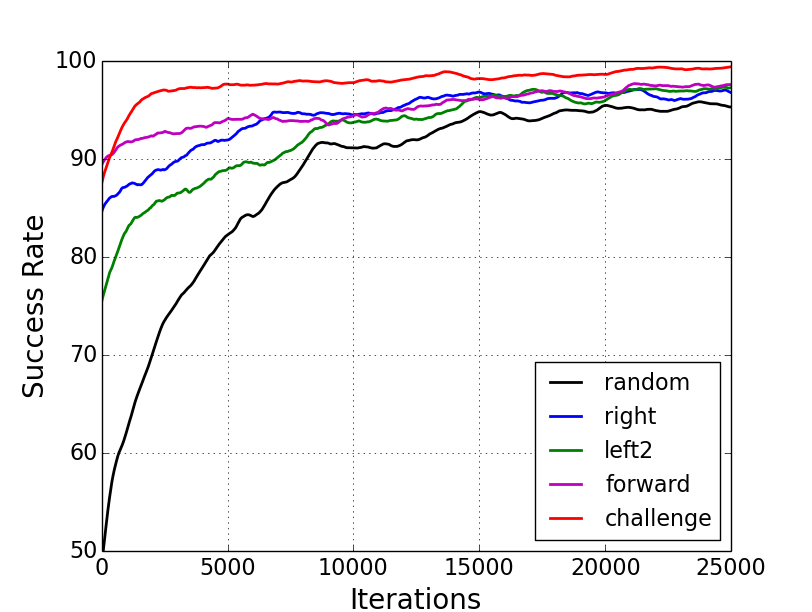}
}
\subfloat[Left2\label{fig:finetune_left2}]{%
\includegraphics[height=.15\textwidth]{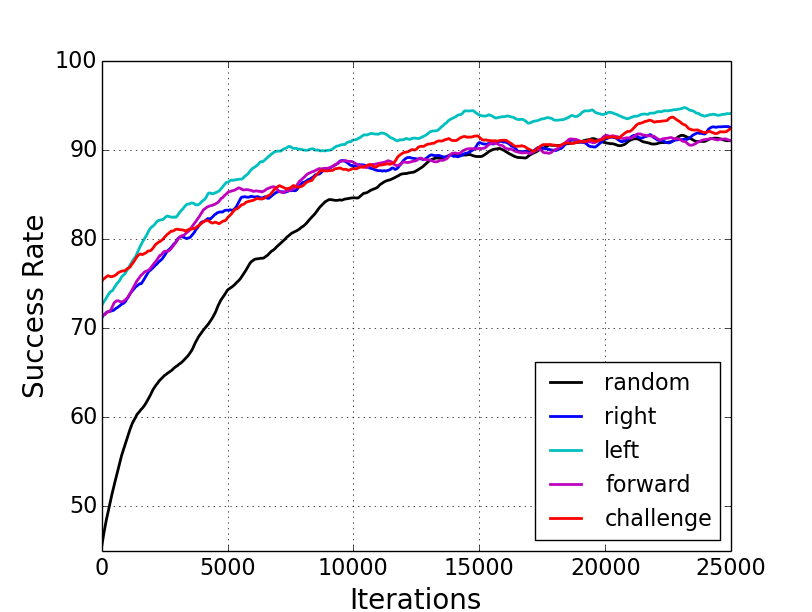}
}
\subfloat[Forward\label{fig:finetune_forward}]{%
\includegraphics[height=.15\textwidth]{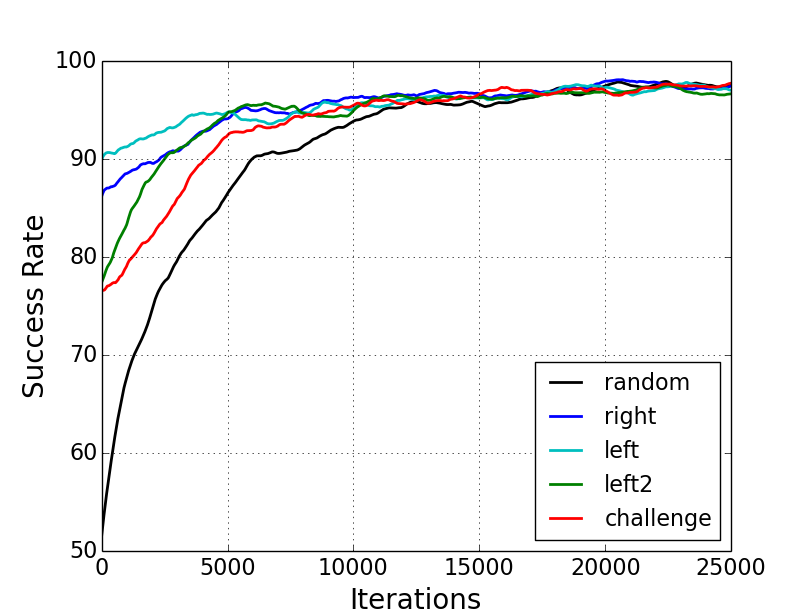}
}
\subfloat[Challenge\label{fig:finetune_challenge}]{%
\includegraphics[height=.15\textwidth]{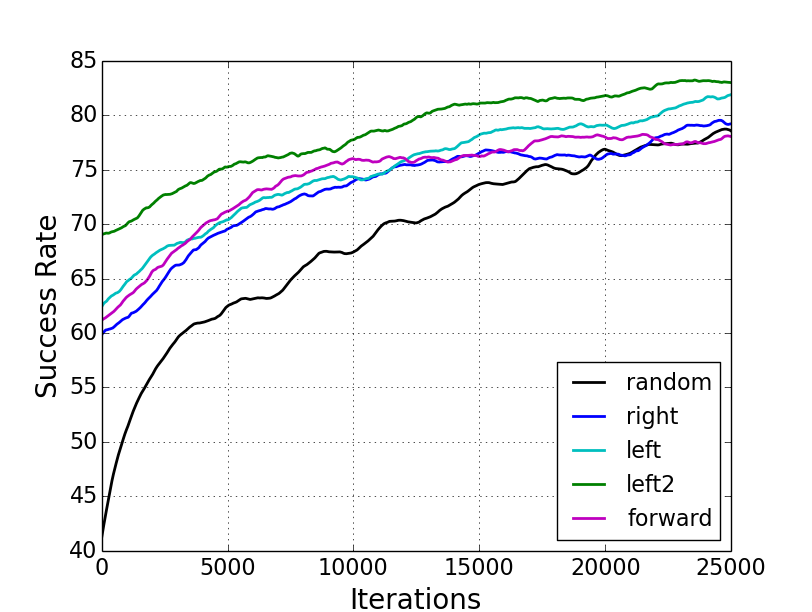}
}
\caption{Fine-tuning comparison. A network for one task is initialized with the network of a different task. The colored lines indicate the initialization network. The black line indicates the performance of a network trained with a random initialization. Initializing a network with a network trained on another task is almost always advantageous. We notice a jumpstart benefit in every tested example, and observe several asymptotic improvements. }\label{fig:fine-tune}
\end{figure*}

\section{Knowledge Transfer}
\label{sec:knowledge}

We investigate the benefits of policy re-use for sharing knowledge between different driving tasks. By sharing knowledge from different tasks we can reduce learning time and create more general and capable systems. Ideally knowledge sharing can be extended to involve a system that continues to learn after it has been deployed \cite{Thrun1996} and can enable a system to accurately predict appropriate behavior in novel situations \cite{isele2016task}. We examine the behavior of various knowledge sharing strategies in the autonomous driving domain.

\subsection{Direct Copy} 
Directly copying a policy indicates the differences between tasks. 
To demonstrate how well a network trained on one task fits another, we train a network on a single source task for 25,000 iterations. The \emph{unmodified} network is then evaluated on every other task. We repeat this process, using each different task as a source task. 

\subsection{Fine-Tuning}
Fine-tuning allows a network to adapt from the source to the target task. Starting with a network trained for 10,000 iterations on a \emph{source} task, we then fine-tune a network for an additional 25,000 iterations on a second \emph{target} task. We use 10,000 iterations because it demonstrates substantial learning, but is suboptimal in order to emphasize the possible benefits gained from transfer. Fine-tuning demonstrates the \emph{jumpstart} and \emph{asymptotic performance} as described by Taylor and Stone (\citeyear{taylor2009transfer}).

\subsection{Reverse Transfer}
After a network has been fine-tuned on the target task, we evaluate the performance of that network on the source task. If training on a later task improves the performance of an earlier task this is known as reverse transfer. It is known that neural networks often \emph{forget} earlier tasks in what has been termed catastrophic forgetting \cite{mccloskey1989catastrophic,ratcliff1990connectionist,goodfellow2013empirical}. Since we are interested in a system learning a large variety of intersections, we wish to understand how much knowledge of a previous task is preserved. 

\section{Experimental Setup}
\label{sec:experiments}

% SIMULATION
Experiments were run using the Sumo simulator \cite{sumo}, which is an open source traffic simulation package. 
Traffic scenarios like multi-lane intersections can be setup by defining the road network (lanes and intersections) along with specifications that control traffic conditions. To simulate traffic, users have control over the types of vehicles, road paths, vehicle density, and departure times. Traffic cars follow the intelligent driver model to control their motion. In Sumo, randomness is simulated by varying the speed distribution of the vehicles and by using parameters that control driver imperfection (based on the Krauss stochastic driving model \cite{krauss1998sumo}). The simulator runs based on a predefined time interval which controls the length of every step. We ran experiments using five different intersection scenarios: \emph{Right}, \emph{Left}, \emph{Left2}, \emph{Forward} and a \emph{Challenge}. Each of these scenarios is depicted in Figure \ref{fig:scenarios}.

The Sumo traffic simulator is configured so that each lane has a 45 miles per hour (20 m/s) max speed. The car begins from a stopped position. Each time step is equal to 0.2 seconds. The max number of steps per trial is capped at 100 steps (20 seconds). The traffic density is set by the probability that a vehicle will be emitted randomly per second. We use depart probability of 0.2 for \emph{each lane} for all tasks.

% metrics
While navigating intersections involves multiple conflicting metrics (including time to cross, number of collisions, and disruption to traffic), we focus on the percentage of trials the vehicle successfully navigates the intersection. All simulated state representations ignore occlusion, assuming all cars are always visible.

\subsection{Deep Neural Network Setup}

Our DQN uses a convolutional neural network with two convolution layers, and one fully connected layer. The first convolutional layer has $32$ $6 \times 6$ filters with stride two, the second convolution layer has $64$ $3 \times 3$ filters with stride two. The fully connected layer has 100 nodes. All layers use leaky ReLU activation functions \cite{maas2013rectifier}. The final linear output layer has five outputs: a single \emph{go} action, and a \emph{wait} action at four time scales (1, 2, 4, and 8 time steps) inspired by dynamic frame skipping techniques \cite{srinivas2017dynamic}. The network is optimized using the RMSProp algorithm \cite{tieleman2012lecture}. 

At each learning iteration we samples a batch of 60 experiences. 
Since the use of an experience replay buffer imposes a delay between an experience occurring and being trained on, we are able to calculate the return for each state-action pair in the trajectory prior to adding each step into the replay buffer. This allows us to train directly on the n-step return \cite{peng1996incremental} and forgo the added complexity of using target networks \cite{mnih2015human}.

The state space of the DQN is represented as a $18 \times 26$ grid in local coordinates that denote the speed and direction of cars within the grid. The epsilon governing random exploration is $0.05$. The reward  is $+1$ for successfully navigating the intersection, $-1$ for a collision, and $-0.01$ step cost. 

% real data
\subsection{Real data}
We conducted a preliminary study in order to evaluate whether knowledge obtained by simulated intersections could be useful for real ones. We collected data from an autonomous vehicle in Mountain View, California, at an unsigned T-junction, similar to the \emph{Left} scenario. A point cloud, obtained by a combination of six IBEO Lidar sensors, is first pre-processed to remove points that reside outside the road boundaries. A clustering method with hand-tuned geometric thresholds is used for vehicle detection. Each vehicle is tracked by a separate particle filter. During data collection, a human observer in the vehicle labeled at times whether making a left turn would be safe or not. Given a random starting point in the recording, the system is able to select \emph{wait} actions that move it ahead in the recording, and a go action which results in either a collision or a success based on the human provided labels. Because data is collected at a higher sampling rate than the simulation step frequency (and therefore the behavior frequency), states are sampled from within the simulation step window, allowing a recording to be inflated into a large number of experiments. Note that this process only gives real sensor readings, and that the system is not able to observe how its behavior affects other drivers. Because the same few recorded scenarios are replayed, we expect training on recorded data in this way will overfit to the recording. 

\begin{figure}[thp!]
\vspace{-10pt}
\centering
\hspace{-10pt}
  \vspace{-10pt}
  \includegraphics[width=.5\textwidth]{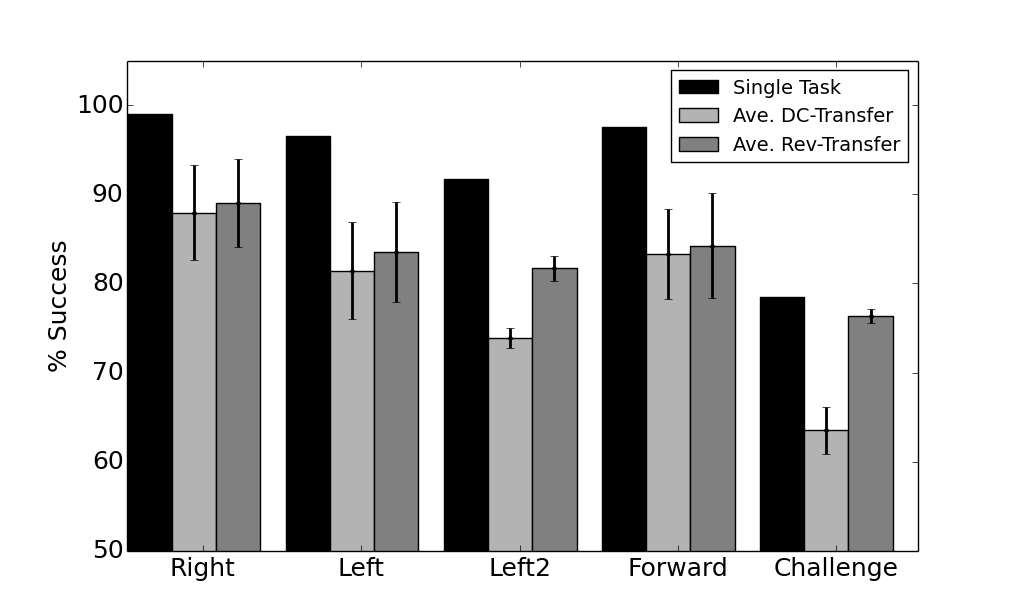}
  \caption{Direct Copy and Reverse Transfer. The x axis denotes the test condition. Black bars show the performance of single task learning. Light gray bars show the average performance of a network trained on one task and tested on another. The drop in performance demonstrates the difference between tasks. The dark gray indicates the average performance of reverse transfer: a network is trained on Task A, fine-tuned on Task B, and then evaluated on Task A. The drop in performance indicates catastrophic forgetting, but networks exhibit some retention of the initial task.}
  \label{fig:dc_rev}
  \vspace{-10pt}
\end{figure}

We train a network on approximately one minute of recorded data during which time approximately 20 cars drive past. We then test the network on a separate recording made at the same intersection. These are preliminary results, as we are currently in the process of collecting a much larger dataset. We compare the results against a network that has been pre-trained   on simulation data and then fine-tuned on the real data. 
% one minute, 20 cars

\section{Results}
\label{sec:results}

\noindent\textbf{Direct Copy:} Figure \ref{fig:dc_rev} shows the average performance of training on one task and applying it to another in light gray. While we only plot the average performance, the quality of transfer is dependent on the particular source and target task. In no instance does a network trained on a different task surpass the performance of a network trained on the matching task, but several tasks achieve similar performance with transfer. Particularly we see that each network trained on a single lane task (right, left, and forward) is consistently a top performer on other single lane tasks. Additionally the more challenging multi-lane settings (left2 and challenge) appear related. The \emph{Left2} network does substantially better than any of the single lane tasks on the \emph{Challenge} task.

\noindent\textbf{Fine-Tuning:} Figure \ref{fig:fine-tune} shows fine-tuning results. We see that in nearly all cases, pre-training with a different network gives a significant advantage in \emph{jumpstart} \cite{taylor2009transfer} and in several cases there is an asymptotic benefit as well. When the fine-tuned networks are re-applied to the source task the performance looks similar to direct copy, as shown in Figure \ref{fig:dc_rev}. 

\noindent\textbf{Reverse Transfer:} The performance on the source task dropped after fine-tuning on the target, but performance improved compared to direct copy. This  indicates that some information was retained by the network. 
Note that the \emph{Left2} and \emph{Challenge} tasks have less overlap with other tasks in the state space. It is possible that non-overlapping regions can be left unchanged by fine-tuning. This might explain why we see the most retention on the least related tasks.  
 
\begin{figure}[t!]
\vspace{-9pt}
\centering
\hspace{-10pt}
	\vspace{-10pt}
  \includegraphics[width=.4\textwidth]{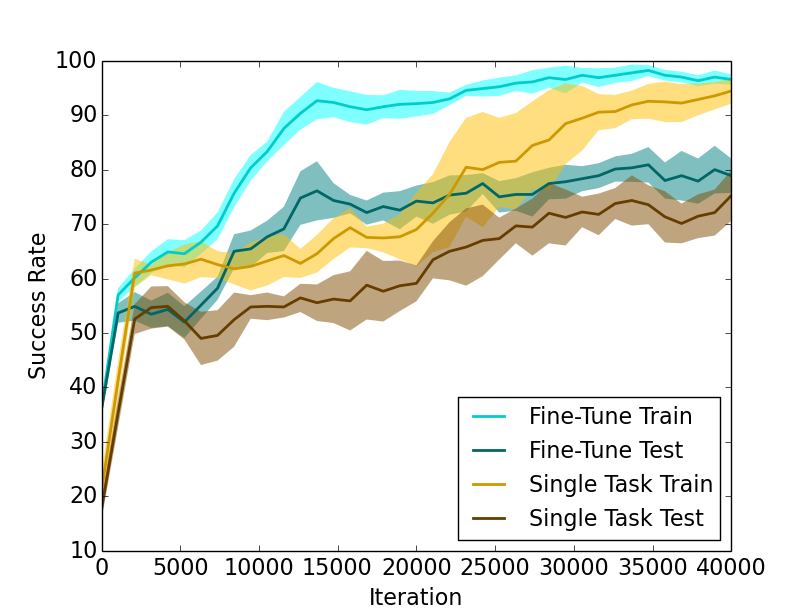}
  \caption{Transfer from Simulation to Real. The yellow lines indicate the training and test performance of a network trained on real data collected from an autonomous vehicle. The blue lines indicate the performance of a network that is first trained on simulated data and then fine-tuned on a real vehicle. We see that fine-tuning speeds up training time and improves generalization. }
  \label{fig:real}
  \vspace{-10pt}
\end{figure}
 
\noindent\textbf{Real Data:} Figure \ref{fig:real} shows the performance of training on real data. The blue lines indicate a network that has been pre-trained on simulated data, the yellow lines indicate the performance of a network that has only been trained on the real data. Lighter lines indicate performance on the training data, and darker lines indicate performance on the test data. The network pre-trained in simulation rises above $90\%$ success on the training data in approximately half the iterations required by the network trained only on real data to reach the same performance. The test results follow the learning curve of the training data, but the performance asymptotes at $80\%$ success. These results show that fine-tuning can reduce the training time of the network, however in both networks the lower performance on the test data suggests over-fitting. 

It is interesting to note that training on the real task required more iterations. This may be due to imperfections in the labeling process or greater noise and variation in the state space. For completeness, we looked at transfer from real data to simulation. We observed that using a network initialized on a real left turn speeds up training a left turn in simulation. Directly applying the network fine-tuned on real data to a left turn in simulation resulted in a model that consistently timed out. We suspect this is due to the network over-fitting the small amount of data, and we will investigate this further when we have collected more data. 
 
\section{Conclusion}
\label{sec:conclusion}

We view autonomous driving as a reinforcement learning problem, and analyze how the knowledge for handling one type of intersection, represented as a Deep Q-Network, translates to other types of intersections. We investigated different properties of transfer between intersections, namely the performance of direct copy, fine-tuning, and reverse transfer and showed how transfer extends from simulated to real intersections. 

Our results identify autonomous intersection handling as a domain that benefits from transfer. First, we found the success rates were consistently low when a network is trained on Task A but directly tested on Task B. Second, a network that is initialized with the network of Task A and then fine-tuned on Task B generally performed better than a randomly initialized network. Third, when a network that is initialized with Task A, fine-tuned on Task B, and tested on Task A, it performed better than a network directly copied from Task B to Task A, but worse than a network trained recently on Task A. Fourth, we show that a real intersection can be treated as a separate task and transfer from simulation can be used to improve learning. Moving forward, we are interested in how transfer can be used to improve the training and robustness of a real system.

\FloatBarrier
\bibliographystyle{icml2017}
\small{
\bibliography{refs}

\begin{thebibliography}{37}
\providecommand{\natexlab}[1]{#1}
\providecommand{\url}[1]{\texttt{#1}}
\expandafter\ifx\csname urlstyle\endcsname\relax
  \providecommand{\doi}[1]{doi: #1}\else
  \providecommand{\doi}{doi: \begingroup \urlstyle{rm}\Url}\fi

\bibitem[Barrett et~al.(2010)Barrett, Taylor, and Stone]{barrett2010transfer}
Barrett, Samuel, Taylor, Matthew~E, and Stone, Peter.
\newblock Transfer learning for reinforcement learning on a physical robot.
\newblock In \emph{Ninth International Conference on Autonomous Agents and
  Multiagent Systems-Adaptive Learning Agents Workshop (AAMAS-ALA)}, 2010.

\bibitem[Bojarski et~al.(2016)Bojarski, Del~Testa, Dworakowski, Firner, Flepp,
  Goyal, Jackel, Monfort, Muller, Zhang, et~al.]{bojarski2016end}
Bojarski, Mariusz, Del~Testa, Davide, Dworakowski, Daniel, Firner, Bernhard,
  Flepp, Beat, Goyal, Prasoon, Jackel, Lawrence~D, Monfort, Mathew, Muller,
  Urs, Zhang, Jiakai, et~al.
\newblock End to end learning for self-driving cars.
\newblock \emph{arXiv preprint arXiv:1604.07316}, 2016.

\bibitem[Bouton et~al.(2017)Bouton, Cosgun, and Kochenderfer]{bouton2017belief}
Bouton, Maxime, Cosgun, Akansel, and Kochenderfer, Mykel~J.
\newblock Belief state planning for navigating urban intersections.
\newblock \emph{IEEE Intelligent Vehicles Symposium (IV)}, 2017.

\bibitem[Brechtel et~al.(2014)Brechtel, Gindele, and
  Dillmann]{brechtel2014probabilistic}
Brechtel, Sebastian, Gindele, Tobias, and Dillmann, R{\"u}diger.
\newblock Probabilistic decision-making under uncertainty for autonomous
  driving using continuous pomdps.
\newblock In \emph{Intelligent Transportation Systems (ITSC), 2014 IEEE 17th
  International Conference on}, pp.\  392--399. IEEE, 2014.

\bibitem[Caruana(1997)]{Caruana1997}
Caruana, Rich.
\newblock {Multitask Learning}.
\newblock \emph{Machine Learning}, 28:\penalty0 41--75, 1997.

\bibitem[Cosgun et~al.(2017)Cosgun, Ma, Chiu, Huang, Demir, Anon, Lian, Tafish,
  and Al-Stouhi]{cosgun2017towards}
Cosgun, A, Ma, L, Chiu, J, Huang, J, Demir, M, Anon, A, Lian, T, Tafish, H, and
  Al-Stouhi, S.
\newblock Towards full automated drive in urban environments: A demonstration
  in gomentum station, california.
\newblock \emph{IEEE Intelligent Vehicles Symposium (IV)}, 2017.

\bibitem[Goodfellow et~al.(2013)Goodfellow, Mirza, Xiao, Courville, and
  Bengio]{goodfellow2013empirical}
Goodfellow, Ian~J, Mirza, Mehdi, Xiao, Da, Courville, Aaron, and Bengio,
  Yoshua.
\newblock An empirical investigation of catastrophic forgetting in
  gradient-based neural networks.
\newblock \emph{arXiv preprint arXiv:1312.6211}, 2013.

\bibitem[Hanna \& Stone(2017)Hanna and Stone]{hanna2017grounded}
Hanna, Josiah~P and Stone, Peter.
\newblock Grounded action transformation for robot learning in simulation.
\newblock In \emph{Proceedings of the 31st AAAI Conference on Artificial
  Intelligence}, 2017.

\bibitem[Isele et~al.(2016)Isele, Rostami, and Eaton]{isele2016task}
Isele, David, Rostami, Mohammad, and Eaton, Eric.
\newblock Using task features for zero-shot knowledge transfer in lifelong
  learning.
\newblock \emph{In Proceedings of the International Joint Conference on
  Artificial Intelligence}, 2016.

\bibitem[Isele et~al.(2017)Isele, Cosgun, Subramanian, and
  Fujimura]{isele2017navigating}
Isele, David, Cosgun, Akansel, Subramanian, Kaushik, and Fujimura, Kikuo.
\newblock Navigating intersections with autonomous vehicles using deep
  reinforcement learning.
\newblock \emph{arXiv preprint arXiv:1705.01196}, 2017.

\bibitem[Jaderberg et~al.(2016)Jaderberg, Mnih, Czarnecki, Schaul, Leibo,
  Silver, and Kavukcuoglu]{jaderberg2016reinforcement}
Jaderberg, Max, Mnih, Volodymyr, Czarnecki, Wojciech~Marian, Schaul, Tom,
  Leibo, Joel~Z, Silver, David, and Kavukcuoglu, Koray.
\newblock Reinforcement learning with unsupervised auxiliary tasks.
\newblock \emph{arXiv preprint arXiv:1611.05397}, 2016.

\bibitem[Kirkpatrick et~al.(2016)Kirkpatrick, Pascanu, Rabinowitz, Veness,
  Desjardins, Rusu, Milan, Quan, Ramalho, Grabska-Barwinska,
  et~al.]{kirkpatrick2016overcoming}
Kirkpatrick, James, Pascanu, Razvan, Rabinowitz, Neil, Veness, Joel,
  Desjardins, Guillaume, Rusu, Andrei~A, Milan, Kieran, Quan, John, Ramalho,
  Tiago, Grabska-Barwinska, Agnieszka, et~al.
\newblock Overcoming catastrophic forgetting in neural networks.
\newblock \emph{arXiv preprint arXiv:1612.00796}, 2016.

\bibitem[Krajzewicz et~al.(2012)Krajzewicz, Erdmann, Behrisch, and
  Bieker]{sumo}
Krajzewicz, Daniel, Erdmann, Jakob, Behrisch, Michael, and Bieker, Laura.
\newblock Recent development and applications of {SUMO}--simulation of urban
  mobility.
\newblock \emph{International Journal on Advances in Systems and Measurements
  (IARIA)}, 5\penalty0 (3--4), 2012.

\bibitem[Krauss(1998)]{krauss1998sumo}
Krauss, Stefan.
\newblock \emph{Microscopic modeling of traffic flow: Investigation of
  collision free vehicle dynamics}.
\newblock PhD thesis, Deutsches Zentrum fuer Luft-und Raumfahrt, 1998.

\bibitem[Kulkarni et~al.(2016)Kulkarni, Narasimhan, Saeedi, and
  Tenenbaum]{kulkarni2016hierarchical}
Kulkarni, Tejas~D, Narasimhan, Karthik, Saeedi, Ardavan, and Tenenbaum, Josh.
\newblock Hierarchical deep reinforcement learning: Integrating temporal
  abstraction and intrinsic motivation.
\newblock In \emph{Advances in Neural Information Processing Systems}, pp.\
  3675--3683, 2016.

\bibitem[Maas et~al.(2013)Maas, Hannun, and Ng]{maas2013rectifier}
Maas, Andrew~L, Hannun, Awni~Y, and Ng, Andrew~Y.
\newblock Rectifier nonlinearities improve neural network acoustic models.
\newblock In \emph{Proc. ICML}, volume~30, 2013.

\bibitem[McCloskey \& Cohen(1989)McCloskey and
  Cohen]{mccloskey1989catastrophic}
McCloskey, Michael and Cohen, Neal~J.
\newblock Catastrophic interference in connectionist networks: The sequential
  learning problem.
\newblock \emph{Psychology of learning and motivation}, 24:\penalty0 109--165,
  1989.

\bibitem[Mnih et~al.(2013)Mnih, Kavukcuoglu, Silver, Graves, Antonoglou,
  Wierstra, and Riedmiller]{mnih2013playing}
Mnih, Volodymyr, Kavukcuoglu, Koray, Silver, David, Graves, Alex, Antonoglou,
  Ioannis, Wierstra, Daan, and Riedmiller, Martin.
\newblock Playing atari with deep reinforcement learning.
\newblock \emph{arXiv preprint arXiv:1312.5602}, 2013.

\bibitem[Mnih et~al.(2015)Mnih, Kavukcuoglu, Silver, Rusu, Veness, Bellemare,
  Graves, Riedmiller, Fidjeland, Ostrovski, et~al.]{mnih2015human}
Mnih, Volodymyr, Kavukcuoglu, Koray, Silver, David, Rusu, Andrei~A, Veness,
  Joel, Bellemare, Marc~G, Graves, Alex, Riedmiller, Martin, Fidjeland,
  Andreas~K, Ostrovski, Georg, et~al.
\newblock Human-level control through deep reinforcement learning.
\newblock \emph{Nature}, 518\penalty0 (7540):\penalty0 529--533, 2015.

\bibitem[Pan \& Yang(2010)Pan and Yang]{Pan2010a}
Pan, Sinno~Jialin and Yang, Qiang.
\newblock {A Survey on Transfer Learning}.
\newblock \emph{IEEE Transactions on Knowledge and Data Engineering},
  22\penalty0 (10), 2010.

\bibitem[Peng \& Williams(1996)Peng and Williams]{peng1996incremental}
Peng, Jing and Williams, Ronald~J.
\newblock Incremental multi-step q-learning.
\newblock \emph{Machine learning}, 22\penalty0 (1-3):\penalty0 283--290, 1996.

\bibitem[Ratcliff(1990)]{ratcliff1990connectionist}
Ratcliff, Roger.
\newblock Connectionist models of recognition memory: Constraints imposed by
  learning and forgetting functions.
\newblock \emph{Psychological review}, 97\penalty0 (2):\penalty0 285--308,
  1990.

\bibitem[Razavian et~al.(2014)Razavian, Azizpour, Sullivan, and
  Carlsson]{Razavian2014}
Razavian, Ali~Sharif, Azizpour, Hossein, Sullivan, Josephine, and Carlsson,
  Stefan.
\newblock {CNN Features off-the-shelf: an Astounding Baseline for Recognition}.
\newblock \emph{Computer Vision and Pattern Recognition Workshops (CVPRW)},
  pp.\  512--519, March 2014.

\bibitem[Rusu et~al.(2016)Rusu, Rabinowitz, Desjardins, Soyer, Kirkpatrick,
  Kavukcuoglu, Pascanu, and Hadsell]{rusu2016progressive}
Rusu, Andrei~A, Rabinowitz, Neil~C, Desjardins, Guillaume, Soyer, Hubert,
  Kirkpatrick, James, Kavukcuoglu, Koray, Pascanu, Razvan, and Hadsell, Raia.
\newblock Progressive neural networks.
\newblock \emph{arXiv preprint arXiv:1606.04671}, 2016.

\bibitem[Shalev-Shwartz et~al.(2016)Shalev-Shwartz, Shammah, and
  Shashua]{shalev2016safe}
Shalev-Shwartz, Shai, Shammah, Shaked, and Shashua, Amnon.
\newblock Safe, multi-agent, reinforcement learning for autonomous driving.
\newblock \emph{arXiv preprint arXiv:1610.03295}, 2016.

\bibitem[Song et~al.(2016)Song, Xiong, and Chen]{song2016intention}
Song, Weilong, Xiong, Guangming, and Chen, Huiyan.
\newblock Intention-aware autonomous driving decision-making in an uncontrolled
  intersection.
\newblock \emph{Mathematical Problems in Engineering}, 2016, 2016.

\bibitem[Srinivas et~al.(2017)Srinivas, Sharma, and
  Ravindran]{srinivas2017dynamic}
Srinivas, Aravind, Sharma, Sahil, and Ravindran, Balaraman.
\newblock Dynamic action repetition for deep reinforcement learning.
\newblock \emph{AAAI Conference on Artificial Intelligence (AAAI)}, 2017.

\bibitem[Srivastava et~al.(2013)Srivastava, Masci, Kazerounian, Gomez, and
  Schmidhuber]{srivastava2013compete}
Srivastava, Rupesh~K, Masci, Jonathan, Kazerounian, Sohrob, Gomez, Faustino,
  and Schmidhuber, J{\"u}rgen.
\newblock Compete to compute.
\newblock In \emph{Advances in neural information processing systems}, pp.\
  2310--2318, 2013.

\bibitem[Sutton \& Barto(1998)Sutton and Barto]{sutton1998reinforcement}
Sutton, Richard~S and Barto, Andrew~G.
\newblock \emph{Reinforcement learning: An introduction}, volume~1.
\newblock MIT press Cambridge, 1998.

\bibitem[Taylor \& Stone(2009)Taylor and Stone]{taylor2009transfer}
Taylor, Matthew~E and Stone, Peter.
\newblock Transfer learning for reinforcement learning domains: A survey.
\newblock \emph{Journal of Machine Learning Research}, 10\penalty0
  (Jul):\penalty0 1633--1685, 2009.

\bibitem[Tessler et~al.(2016)Tessler, Givony, Zahavy, Mankowitz, and
  Mannor]{tessler2016deep}
Tessler, Chen, Givony, Shahar, Zahavy, Tom, Mankowitz, Daniel~J, and Mannor,
  Shie.
\newblock A deep hierarchical approach to lifelong learning in minecraft.
\newblock \emph{arXiv preprint arXiv:1604.07255}, 2016.

\bibitem[Thrun(1996)]{Thrun1996}
Thrun, Sebastian.
\newblock {Is learning the n-th thing any easier than learning the first?}
\newblock \emph{Advances in neural information processing systems}, pp.\
  640--646, 1996.
\newblock ISSN 1049-5258.
\newblock \doi{10.1.1.44.2898}.

\bibitem[Tieleman \& Hinton(2012)Tieleman and Hinton]{tieleman2012lecture}
Tieleman, Tijmen and Hinton, G.
\newblock Lecture 6.5-rmsprop, coursera: Neural networks for machine learning.
\newblock \emph{University of Toronto, Tech. Rep}, 2012.

\bibitem[Tobin et~al.(2017)Tobin, Fong, Ray, Schneider, Zaremba, and
  Abbeel]{tobin2017domain}
Tobin, Josh, Fong, Rachel, Ray, Alex, Schneider, Jonas, Zaremba, Wojciech, and
  Abbeel, Pieter.
\newblock Domain randomization for transferring deep neural networks from
  simulation to the real world.
\newblock \emph{arXiv preprint arXiv:1703.06907}, 2017.

\bibitem[Watkins \& Dayan(1992)Watkins and Dayan]{watkins1992q}
Watkins, Christopher~JCH and Dayan, Peter.
\newblock Q-learning.
\newblock \emph{Machine learning}, 8\penalty0 (3-4):\penalty0 279--292, 1992.

\bibitem[Yin \& Pan(2017)Yin and Pan]{yin2017knowledge}
Yin, Haiyan and Pan, Sinno~Jialin.
\newblock Knowledge transfer for deep reinforcement learning with hierarchical
  experience replay.
\newblock In \emph{AAAI Conference on Artificial Intelligence (AAAI)}, 2017.

\bibitem[Yosinski et~al.(2014)Yosinski, Clune, Bengio, and
  Lipson]{Yosinski2014}
Yosinski, Jason, Clune, Jeff, Bengio, Yoshua, and Lipson, Hod.
\newblock {How transferable are features in deep neural networks ?}
\newblock \emph{NIPS}, 27, 2014.

\end{thebibliography}
}

\end{document}